\documentclass{article}
\usepackage{spconf,amsmath,graphicx}

\usepackage{bm}
\usepackage{arydshln}
\usepackage{multirow}
\usepackage{booktabs}
\usepackage[bb=boondox]{mathalfa}

\title{In-Context Learning for Few-Shot Nested Named Entity Recognition}

\name{Meishan Zhang$^1$, Bin Wang$^1$, Hao Fei$^2$\sthanks{Corresponding author: Hao Fei}, Min Zhang$^1$}
\address{$^1$ Harbin Institute of Technology (Shenzhen), Shenzhen, China\\
$^2$ National University of Singapore, Singapore
}

\begin{document}

\maketitle

\begin{abstract}
In nested Named entity recognition (NER), entities are nested with each other, and thus requiring more data annotations to address.
This leads to the development of few-shot nested NER, where the prevalence of pretrained language models with in-context learning (ICL) offers promising solutions.
In this work, we introduce an effective and innovative ICL framework for the setting of few-shot nested NER. 
We improve the ICL prompt by devising a novel example demonstration selection mechanism, \textbf{EnDe} retriever. 
In EnDe retriever, we employ contrastive learning to perform three types of representation learning, in terms of semantic similarity, boundary similarity, and label similarity, to generate high-quality demonstration examples. 
Extensive experiments over three nested NER and four flat NER datasets demonstrate the efficacy of our system.
\end{abstract}


\section{Introduction}
\label{sec:intro}
\begin{figure}[!t]
  \centering
  \centerline{\includegraphics[width=0.98\linewidth]{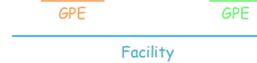}}
% \vspace{-2mm}
\caption{Illustration of nested NER.}
% \vspace{-5mm}
\label{fig:intro}
\end{figure}

NER is a fundamental task in natural language processing (NLP) that identifies and classifies entities in unstructured text \cite{tjong-kim-sang-de-meulder-2003-introduction}. 
The evolution of NER has led to the development of nested NER \cite{finkel-manning-2009-nested} from the flat NER, a more intricate variant, where entities can be nested with each other \cite{fei2020boundaries}. 
Unlike regular NER, nested NER introduces complexity by allowing multiple entity types to coexist within the same entity token, as exemplified in Fig. \ref{fig:intro}.
Researchers have explored various approaches to tackle nested NER, including complex sequential labeling methods \cite{ju-etal-2018-neural}, graph-based methods \cite{lu-roth-2015-joint}, and generative methods \cite{yan-etal-2021-unified-generative,0001WLLLQZZC22}. 
It has been extensively identified by existing research that the key to nested NER lies in accurately delineating the boundaries of nested entities within the text, i.e., when an entity begins and ends \cite{0001JLLRL21}.

Few-shot NER has become pivotal research due to the challenges posed by the scarcity of labeled data for training NER models, especially for the nested scenario.
Recent years have witnessed the remarkable development of pretrained Language Models (LMs) such as T5 \cite{RaffelSRLNMZLL20} and GPT-3.5\footnote{https://platform.openai.com/docs/guides/gpt}. 
LMs have demonstrated exceptional few-shot learning capabilities \cite{wu2023nextgpt}, making them promising for tackling few-shot NER \cite{abs-2012-14978}. 
Leveraging the power of LMs as backbones, few-shot NER involves providing LMs with a few example entities and prompting them to predict all possible mentions and labels for a given sentence \cite{chen-etal-2022-shot}.
The introduction of in-context learning (ICL) \cite{BrownMRSKDNSSAA20} has further enhanced the efficacy of LM, where LMs are provided with label demonstrations and semantic context for stronger few-shot NER capabilities.

It's worth noting that, as revealed by existing findings \cite{min-etal-2022-rethinking}, the effectiveness of ICL-based few-shot NER heavily relies on the selection of appropriate demonstration examples. 
Prior research has demonstrated that the choice of appropriate ICL examples plays a critical role in prompting LMs to produce accurate and informative NER predictions.
This principle equally applies to few-shot nested NER, where selecting the optimal ICL examples becomes even more challenging due to the complex entity boundaries.
We observe at least three key challenges:
\textbf{First}, the selected ICL examples should have high semantic similarity to the input test sentence. 
Often, sentences with more identical semantics involve similar entities and similar labels.
\textbf{Second}, the entities within the chosen ICL examples should align closely with the boundaries of entities within the test instance. 
This alignment is particularly crucial for nested NER, as it relies heavily on boundary detection.
\textbf{Third}, for nested entities, the labels should be carefully chosen to differentiate between entities that may share similar word semantic representations. 
Even when two nested entities have close word representations, their labels should have sufficient differentiation to avoid ambiguity.

\begin{figure}[!t]
  \centering
  \centerline{\includegraphics[width=0.98\linewidth]{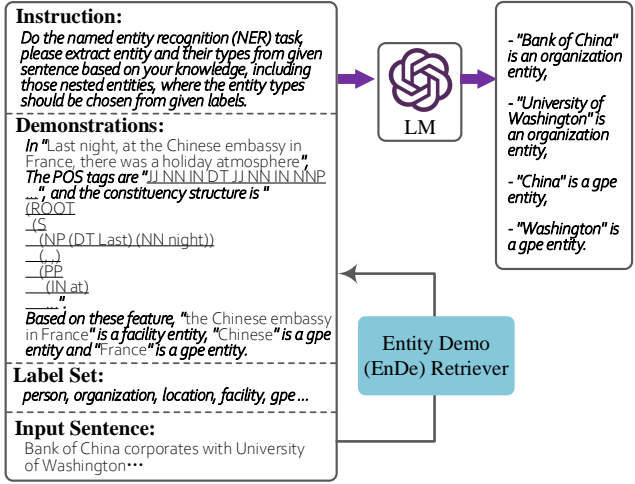}}
% \vspace{-2mm}
\caption{Illustration of the prompts with in-context learning.
}
% \vspace{-3mm}
\label{fig:icl}
\end{figure}

Based on the above observations, this work introduces an innovative ICL framework for highly effective few-shot nested NER.
In our approach, we first devise an integral ICL prompt template, as illustrated in Fig. \ref{fig:icl}, which comprises four main components: \textbf{Task Instruction}, \textbf{Demonstrations}, \textbf{Label Set} and \textbf{Testing Sentence}. 
The heart of our methodology lies in the meticulous selection of demonstrations.
For each demonstration example, besides the entity labels, we also enable additional descriptions of NER entity boundaries, including the linguistic part-of-speech (POS) tags, and the syntactic constituency structure of the sentence.
These features have been highlighted by prior research to be effective in accurately defining mention boundaries \cite{fei-etal-2021-better}.

To identify the most effective examples and offer optimal guidance for the test instance, we devise a novel Entity Demo (\textbf{EnDe}) Retriever mechanism, which considers two key aspects during the selection process:
a) EnDe selects examples with higher comparable sentence semantics to provide better guidance;
b) examples with higher degrees of boundary similarity with the test instance are chosen.
To realize these measurements, we employ contrastive learning \cite{chopra2005learning} for enhanced representation learning, which brings sentence representations with higher semantic and boundary similarity closer in feature space, thus improving few-shot NER performance.
Furthermore, we delve deeper into the EnDe Retriever by incorporating a label-centric representation learning, which reduces the feature distance between ICL examples with the same entity label, and meanwhile increases the distance between examples with different labels but token overlap.
The experimental results unequivocally demonstrate that our approach consistently achieves new state-of-the-art performance in few-shot NER across three nested NER and four flat NER benchmark datasets, validating the robustness and efficacy of our framework.

\begin{figure}[!t]
  \centering
  \centerline{\includegraphics[width=0.98\linewidth]{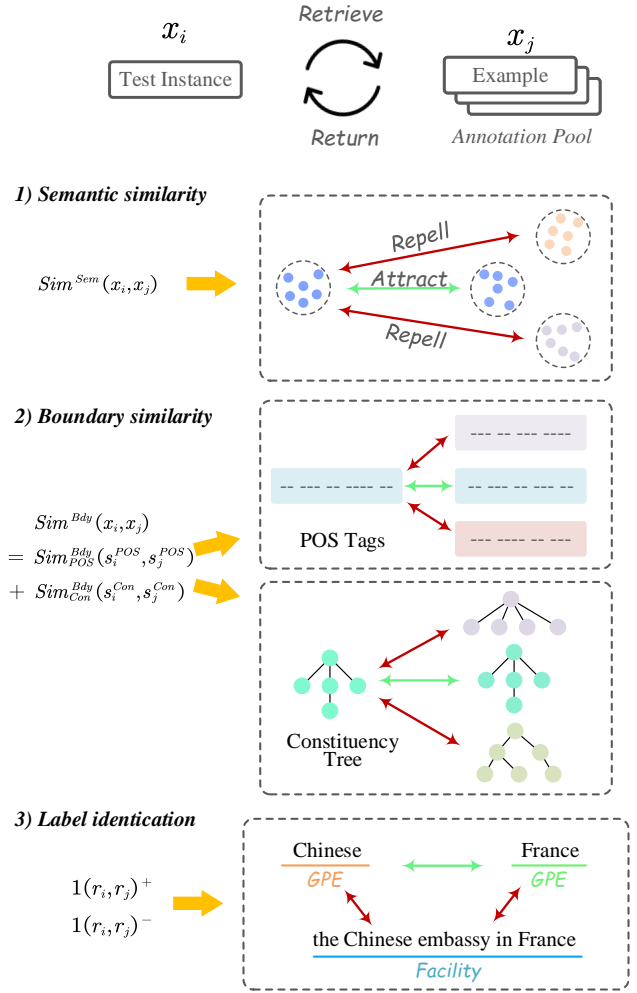}}
% \vspace{-3mm}
\caption{The framework of EnDe Retriever.}
% \vspace{-4mm}
\label{fig:EnDe}
\end{figure}

\section{Methodology}
\label{sssec:subsubhead}

% \vspace{-2mm}
\subsection{Task Definition}

% \vspace{-2mm}
Formally, we frame the NER task as a problem of classifying entity spans to encompass all possible word combinations. 
To be specific, given an input sentence $x$ consisting of $n$ words, $x = \{w_1, \cdots , w_n\}$, we generate a set of entity spans, $\{e_i\}$ that consists of a sequence of words $e_i=\{w_p, \cdots , w_q\}$ (where $1 \le p \le q \le n$) with a specific mention type label $r_i \in R$ ($R$ is a pre-defined label set), where $e_i$ can be overlapped with $e_j$ on certain tokens.
For the few-shot nested NER, only a few numbers ($k$-shot) of annotated examples are used for the model to learn from.

\begin{table*}[!ht]
\centering
\fontsize{9}{12}\selectfont
\setlength{\tabcolsep}{2.5mm}
\begin{tabular}{lccccccccc}
\hline
 & \multicolumn{3}{c}{ACE2004		} & \multicolumn{3}{c}{ACE2005		} & \multicolumn{3}{c}{GENIA		} \\ 
\cmidrule(r){2-4} \cmidrule(r){5-7} \cmidrule(r){8-10}
& 5-shot& 10-shot& 20-shot&  5-shot& 	10-shot& 	20-shot&  	5-shot& 	10-shot& 	20-shot	 \\ \hline
% \multirow{6}{*}{\textbf{\em Syntax-free Approach}} 
Loc-Lab (Shen et al., 2021) \cite{shen-etal-2021-locate} &		7.20 &		22.88 &		41.02 &		11.43	 &	25.13 &		39.61 &		15.57 &		31.65 &		49.67 \\
GNER (Yan et al., 2021) \cite{yan-etal-2021-unified-generative} &		8.87 &		14.19	 &	28.73	 &	8.72 &		13.17 &		24.26 &		4.68 &		10.62 &		20.98 \\
SEE (Yang et al., 2022) \cite{yang-etal-2022-see} &		26.54 &		38.89	 &	48.94	 &	25.58	 &	36.36 &		51.31 &		19.31 &		37.78 &		50.93 \\
SDNet (Chen et al., 2022) \cite{chen-etal-2022-shot} &		20.55 &		34.82	 &	42.87 &		22.03	 &	32.20 &		43.00 &		17.46 &		19.03 &		33.27 \\
ESD (Wang et al., 2022b) \cite{wang-etal-2022-enhanced} &		19.25 &		42.75 &		52.17 &		31.57 &		38.81 &		50.30 &		25.03 &		35.23 &		47.22 \\
W2NER (Li et al., 2022) \cite{li2022unified} &		12.52 &		35.84 &		45.73 &		15.48 &		37.41	 &	43.54 &		14.25 &		20.42 &		34.84 \\
FIT (Xu et al., 2023) \cite{xu-etal-2023-focusing} &		35.87 &		44.88 &		53.92 &		37.74	 &	42.25	 &	52.71 &		34.43 &		44.95 &		51.26 \\

 \cdashline{1-10}
Ours (T5-base) & \bf	40.15 & \bf	48.43 & \bf	55.45 & \bf	44.42 & \bf	46.10 & \bf	54.98 & \bf	39.85 & \bf	48.27 & \bf	54.39 \\

\hline
\end{tabular}
% \vspace{-1mm}
\caption{Main results (F1) on three datasets under different shot sizes.
}
\vspace{-3mm}
\label{tab:main}
\end{table*}

\subsection{Generative NER with In-context Learning}

Based on the generative LM backbone, e.g., T5 or GPT-3.5, we transform NER into a text-to-text prediction.
We combine the In-context learning (ICL), which as demonstrated in prior studies (Brown et al., 2020; Lee et al., 2022b), further improves the model's few-shot capabilities, particularly in NER tasks.
As shown in Fig. \ref{fig:icl}, the prompt comprises four parts:

\textbf{Task Instruction} is a prompt that instructs the model about the current task it should perform. 
The instruction for the main NER task is: \emph{extracting entity and their types from a given sentence based on your knowledge}.

\textbf{Demonstrations} offer input and output examples in demonstrative formats.
We also explicitly mark the boundary features to offer more features.

\textbf{Label Set} is a list of entity types $R$, from which the model needs to select the corresponding label to annotate the corresponding span. 
The label serves as a constraint and guidance for entity type selection.

\textbf{Testing Sentence} is the input $x$ from which entities need to be extracted by LM.

% \vspace{-4mm}
\subsection{Entity Demonstration (EnDe) Retriever}

\vspace{-2mm}
As high-quality examples are pivotal to ICL for better prompting LM to induce more correct results, we propose an EnDe retriever.
As illustrated in Fig. \ref{fig:EnDe}, EnDe retriever performs representation learning based on contrastive learning \cite{chopra2005learning} under three types of similarity measurements: semantic similarity, boundary similarity and label similarity.

% \vspace{-3mm}
\subsubsection{Semantic Similarity Measurement}

% \vspace{-2mm}
Given a test sentence $x_i$, EnDe first retrieves from the annotation pool (training set) examples with higher sentence semantics.
Thus, we define the contrastive loss as:
\begin{equation}\small
\setlength\abovedisplayskip{2pt}
\setlength\belowdisplayskip{2pt}
\mathcal{L}^{\text{Sem}} = - \sum_i^{K}  \sum_{i\in Q_i} \log \frac{
\exp{[\text{Sim}^{\text{Sem}}( \bm{x}_i  ||  \bm{x}_j )}
]}{
\sum_{*\in N(i,j)} \exp{[\text{Sim}^{\text{Sem}}( \bm{x}_i || \bm{x}_{*} )]}
} \,,
\end{equation}
where 
$Q_i$ contains all the positive examples that have high similarity with $x_i$, i.e., $\text{Sim}^{\text{Sem}}(a || b) > 0.5$ which is a cosine function.
$N(i,j)$ contains set of negative pairs.

% \vspace{-3mm}

\subsubsection{Boundary Similarity Measurement}
% \label{sec:copyright}

% \vspace{-1mm}
Boundary features are essential to nested NER, where here we consider the POS tags and the constituency trees.
Also, examples with higher degrees of boundary similarity with the test instance are chosen.
We use LSTM model to encode the POS tag sequence of a sentence, into $s^{POS}_i=$LSTM($x_i^{POS}$), and likewise, use GCN to encode the constituency tree into representation $s^{Con}_i=$GCN($x_i^{Con}$).
We then define the contrastive loss as:
\begin{equation}\small
\begin{aligned}
\setlength\abovedisplayskip{2pt}
\setlength\belowdisplayskip{2pt}
\mathcal{L}^{\text{Bdy}} =&  \mathcal{L}^{\text{Bdy}}_{\text{POS}} + \mathcal{L}^{\text{Bdy}}_{\text{Con}} \\
=& - \sum_i^{K} \sum_{i\in Q_i} \log \frac{
\exp{[\text{Sim}^{\text{Bdy}}_{\text{POS}}( \bm{s}^{\text{Con}}_i  ||  \bm{s}^{\text{Con}}_j )}
]}{
\sum_{*\in N(i,j)} \exp{[\text{Sim}^{\text{Bdy}}_{\text{POS}}( \bm{s}^{\text{Con}}_i || \bm{s}^{\text{Con}}_{*} )]}
} \\
&- \sum_i^{K} \sum_{i\in Q_i} \log \frac{
\exp{[\text{Sim}^{\text{Bdy}}_{\text{Con}}( \bm{s}^{\text{Con}}_i  ||  \bm{s}^{\text{Con}}_j )}
]}{
\sum_{*\in N(i,j)} \exp{[\text{Sim}^{\text{Bdy}}_{\text{Con}}( \bm{s}^{\text{Con}}_i || \bm{s}^{\text{Con}}_{*} )]}
} \,.
\end{aligned}
\end{equation}
 % \frac{1}{|Q_i|}

% \vspace{-3mm}
\subsubsection{Label Identication Measurement}
% \label{sec:copyright}

% \vspace{-1mm}
It is also important in nested NER that the overlapped tokens of two nested entities have shared word representations.
However, the two entities may largely have distinct mention types, i.e., different label semantics.
Thus, we further propose a label-centric representation learning, where the contrastive loss is defined as:
\begin{equation}\small
\setlength\abovedisplayskip{2pt}
\setlength\belowdisplayskip{2pt}
\mathcal{L}^{\text{Lab}} = - \sum_i^{K}  \log \frac{
\exp{[\mathbb{1}( {r}_i ,  {r}_j )^{+}}
]}{
\sum_{*\in N(i,j)} \exp{[\mathbb{1}({r}_i , {r}_*)^{+} + \mathbb{1}({r}_i ,  {r}_* )^{-}}
]} \,,
\end{equation}
where we put semantically closer those entities in the same labels, $({r}_i ,  \bm{r}_j)^{+}$.

% \vspace{-2mm}
\section{Experiments}
\label{sec:refs}

% \vspace{-2mm}
\subsection{Datasets and Settings}

% \vspace{-1mm}
In line with prior research, we employ standard NER benchmarks, including three nested NER datasets \cite{xu-etal-2023-focusing} (ACE2004, ACE2005, GENIA) and four flat NER datasets \cite{chen-etal-2022-shot} (KBP2017, CoNLL, WNUT, OntoNotes5). 
Our experiments follow the 5/10/20-shot settings as utilized in previous studies. 
For each dataset, we execute ten runs and report the average F1 score. 
Our baselines consist of strong-performing and state-of-the-art few-shot NER systems (all with BERT-base \cite{devlin-etal-2019-bert} or T5-base \cite{RaffelSRLNMZLL20}), as listed in Table \ref{tab:main}. 
For fair comparisons, our backbone is the generative LM, primarily the T5-base model, which remains frozen, while updates are confined to the EnDe retriever. 
Additionally, we explore LM backbones of varying sizes and types.
We check the F1 metric on the test set, considering a predicted entity as correct if both its entity type and offsets match the gold entity.

\vspace{-2mm}
\subsection{Results on Few-shot Nested NER}

% \vspace{-1mm}
Table \ref{tab:main} presents the performance of different methods on three nested datasets.
Overall, we find that different few-shot methods exhibit varied performance increases as the shot size increases. 
Notably, models perform significantly better in the 20-shot setting compared to the 5-shot setting, as increased examples provide more supervision for learning. 
Crucially, our model consistently surpasses all baselines by a substantial margin. 
Remarkably, our model's improvements become even more pronounced as the shot sizes decrease. 
In the 5-shot setting, our model's F1-scores outperform the second-best model by large margins.
This underscores the effectiveness of our approach in addressing nested NER tasks with limited supervision.

\begin{table}[!t]
\centering
\fontsize{9}{12}\selectfont
\setlength{\tabcolsep}{0.9mm}
\begin{tabular}{l ccccc}
\hline	
 & KBP2017 & 	CoNLL & 	WNUT & 	Onto & 	\emph{Avg} \\
 \hline
 Proto (Huang et al., 2020) & 	17.3 & 	58.4 & 	29.5 & 	53.3 & 	39.6\\
GNER (Yan et al., 2021) & 8.4 & 	55.2 & 	24.8 & 	50.4 & 	34.7\\
SEE (Yang et al., 2022) & 	22.8 & 	67.6 & 	36.2 & 	67.1 & 	48.4\\
W2NER (Li et al., 2022) & 	18.0 & 	54.5 & 	26.7 & 	48.4 & 	36.9\\
SDNet (Chen et al., 2022) & 	20.2 & 	71.4 & 	44.1 & 	71.0 & 	51.6\\
Ours (T5-base)	 & \bf 34.7 & 	\bf 73.1 & 	\bf 48.3	 & \bf 74.7 & 	\bf 57.7\\

\hline
\end{tabular}
% \vspace{-1mm}
\caption{
5-shot results on flat NER.
}
% \vspace{-3mm}
\label{tab:flat}
\end{table}

\begin{figure}[!t]
  \centering
  \centerline{\includegraphics[width=0.98\linewidth]{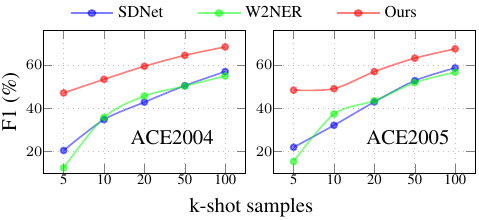}}
  \vspace{-3mm}
\caption{Results with k-shot samples on two datasets.}
% \vspace{-3mm}
\label{fig:Shot}
\end{figure}

% \vspace{-3mm}
\subsection{Results on Few-shot Flat NER}

% \vspace{-1mm}
In Fig. \ref{tab:flat} we further present the performances on the regular flat NER under few-shot learning.
We mainly consider the 5-shot NER.
As seen, our system still outperforms all baselines with clear margins over all datasets, with average 6.1\% F1 improvement, where especially ours surpasses the second-best SEE baseline with 11.9\% F1 on KBP2017 data.
This further indicates that our method's advance can be compatible to the general few-shot NER setting.

% \vspace{-3mm}
\subsection{Effects of Shot Size}

% \vspace{-1mm}
To assess our system's performance across various shot settings, we evaluate it alongside two robust baselines, considering k-shot samples ranging from 5 to 100. 
Intuitively, more shot sizes allow better overall performances.
As depicted in Fig. \ref{fig:Shot}, our system consistently outperforms all others across different shot settings. 
Notably, the enhancements are more pronounced in low-shot settings, affirming the underlying principles of our approach.
Further, we find that for our system, purely increasing the number of examples leads to limited performance enhancement.
This is largely because our EnDe retriever helps already collect the most informative examples, from which the LM benefits the most.

\begin{figure}[!t]
  \centering
  \centerline{\includegraphics[width=0.98\linewidth]{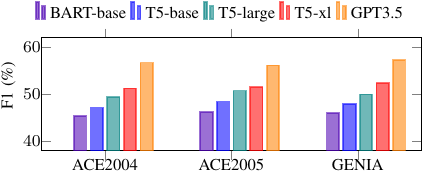}}
  \vspace{-3mm}
\caption{Performances of our system by employing LMs in different types and sizes.}
% \vspace{-3mm}
\label{fig:LMs}
\end{figure}

% \vspace{-3mm}
\subsection{Impacts of Using different LMs}

% \vspace{-1mm}
Finally, we study the influences of adopting different sizes and types of backbone LMs.
We experiment on three datasets, with five different LMs, including BART-base, T5-base, T5-large, T5-xl, and GPT-3.5.
Fig. \ref{fig:LMs} plots the patterns.
We can find that with the increase of LM sizes, i.e., from T5-base to T5-xl, the performances grow steadily.
When using the current state-of-the-art GPT-3.5 LM (with 175B), we find there is a huge performance leap consistently.
This is intuitive as larger LM shows better unsupervised inference capability.

% \vspace{-2mm}
\section{Conclusion}

% \vspace{-2mm}

This paper introduces an innovative in-context learning framework to address the less-explored few-shot nested NER task. 
We improve the ICL prompt by introducing an efficient example demonstration selection mechanism known as the EnDe retriever. 
In the EnDe retriever, we employ contrastive learning to perform three types of representation learning measurements, which include semantic similarity, boundary similarity, and label similarity, to generate high-quality demonstration examples. 
Comprehensive experiments conducted on three nested NER and four flat NER datasets confirm the effectiveness of our system.

\section{Acknowledgement}

This work is supported by the National Natural Science Foundation of China (Grant No. 62176180), and CCF-Baidu Open Fund.

\newpage

\bibliographystyle{IEEEbib}
\bibliography{strings,refs}

\end{document}